\documentclass{article}
\usepackage{iclr2026_conference,times}

\usepackage{amsmath,amsfonts,bm}

\def\eqref#1{equation~\ref{#1}}

\def\1{\bm{1}}

\DeclareMathAlphabet{\mathsfit}{\encodingdefault}{\sfdefault}{m}{sl}
\SetMathAlphabet{\mathsfit}{bold}{\encodingdefault}{\sfdefault}{bx}{n}

\usepackage{hyperref}
\usepackage{url}
\usepackage{amsmath,amssymb,amsfonts}
\usepackage{algorithmic}
\usepackage{algorithm}
\usepackage{graphicx}
\usepackage{booktabs}
\usepackage{multirow}

\iclrfinalcopy

\title{Adaptive Memory Admission Control for LLM Agents}

\author{
\textbf{Guilin Zhang}\thanks{Corresponding authors: Guilin Zhang and Kai Zhao.} \quad
\textbf{Wei Jiang} \quad
\textbf{Xiejiashan Wang} \quad
\textbf{Aisha Behr} \\
\textbf{Kai Zhao}$^*$ \quad
\textbf{Jeffrey Friedman} \quad
\textbf{Xu Chu} \quad
\textbf{Amine Anoun} \\[2pt]
\textbf{Workday AI}
}

\begin{document}

\maketitle
\thispagestyle{fancy}
\lhead{Published at the ICLR 2026 Workshop MemAgent}

\begin{abstract}
LLM-based agents increasingly rely on long-term memory to support multi-session reasoning and interaction, yet current systems provide little control over what information is retained. In practice, agents either accumulate large volumes of conversational content, including hallucinated or obsolete facts, or depend on opaque, fully LLM-driven memory policies that are costly and difficult to audit. As a result, memory admission remains a poorly specified and weakly controlled component in agent architectures. To address this gap, we propose Adaptive Memory Admission Control (A-MAC), a framework that treats memory admission as a structured decision problem. A-MAC decomposes memory value into five complementary and interpretable factors: future utility, factual confidence, semantic novelty, temporal recency, and content type prior. The framework combines lightweight rule-based feature extraction with a single LLM-assisted utility assessment, and learns domain-adaptive admission policies through cross-validated optimization. This design enables transparent and efficient control over long-term memory. Experiments on the LoCoMo benchmark show that A-MAC achieves a superior precision–recall tradeoff, improving F1 to 0.583 while reducing latency by 31\% compared to state-of-the-art LLM-native memory systems. Ablation results identify content type prior as the most influential factor for reliable memory admission. These findings demonstrate that explicit and interpretable admission control is a critical design principle for scalable and reliable memory in LLM-based agents. Code is available at \url{https://github.com/GuilinDev/Adaptive_Memory_Admission_Control_LLM_Agents}.
\end{abstract}

\section{Introduction}

Large language model (LLM) agents have demonstrated strong capabilities in multi-turn interaction, reasoning, and tool use~\citep{generative_agents,voyager,reflexion,chain_of_thought,react,toolformer}. As these agents increasingly operate across extended interactions, long-term memory becomes a core architectural component for maintaining information beyond the context window and supporting coherent behavior over time~\citep{memory_survey_2024}. However, determining what information should be retained in long-term memory remains a critical challenge. Indiscriminate storage leads to bloated memory stores and increased retrieval latency~\citep{rag,hipporag}, while the retention of hallucinated or outdated information can propagate errors into future interactions~\citep{hallucination_survey}. At the other extreme, overly conservative admission policies risk discarding information that is essential for task continuation and long-horizon reasoning.

Existing memory management approaches fall broadly into two categories, each with notable limitations. Heuristic-based methods such as MemGPT~\citep{memgpt} and MemoryBank~\citep{memorybank} rely on hand-crafted scoring functions that combine factors such as recency, relevance, and importance. While computationally efficient, these approaches lack principled mechanisms for preventing hallucinated content from entering memory and struggle with subtle admission decisions. In contrast, LLM-native approaches such as A-mem~\citep{amem} and Mem0~\citep{mem0} delegate memory admission entirely to language models by generating structured memory attributes. Although effective in recall, these methods incur substantial computational overhead and offer limited interpretability, making memory policies difficult to audit or debug. Notably, neither class of methods explicitly addresses hallucination as a first-class concern, despite its recognition as a fundamental challenge for agent reliability~\citep{hallucination_survey,constitutional_ai}.

To address these limitations, we propose \textbf{Adaptive Memory Admission Control (A-MAC)}, a framework that treats memory admission as a structured decision problem rather than an implicit byproduct of generation. Instead of relying on ad hoc heuristics or opaque LLM judgments, A-MAC explicitly evaluates candidate memories before they enter long-term storage. This formulation elevates memory admission to a first-class control mechanism in agent architectures, enabling deliberate tradeoffs between memory coverage, reliability, and efficiency.

A-MAC decomposes memory value into five complementary and interpretable dimensions: future utility, factual confidence, semantic novelty, temporal recency, and content type prior. These dimensions capture distinct aspects of memory quality that are often conflated or left implicit in existing approaches. Utility estimates the potential relevance of a memory for future tasks, while confidence assesses whether the information is supported by prior conversational evidence, directly mitigating hallucination propagation. Novelty prevents redundant storage, recency accounts for temporal decay, and content type prior encodes domain knowledge about which categories of information warrant long-term persistence.

The framework combines lightweight rule-based feature extraction with a single LLM-assisted utility assessment, achieving a favorable balance between computational efficiency and semantic expressiveness. Admission policies are learned through cross-validated optimization, allowing A-MAC to adapt to different conversational domains without manual tuning. By explicitly modeling reliability, redundancy, temporal relevance, and content persistence, A-MAC enables transparent, data-driven control over what information is promoted to long-term memory, offering a scalable alternative to fully LLM-native memory systems.

This paper makes the following contributions:
\begin{enumerate}
    \item We identify memory admission as a critical but under-specified control problem in LLM-based agents and analyze the limitations of existing heuristic and fully LLM-driven approaches.
    \item We introduce Adaptive Memory Admission Control (A-MAC), an interpretable framework that evaluates candidate memories using five complementary dimensions that jointly capture value, reliability, and persistence.
    \item We propose an efficient hybrid design that combines rule-based feature computation with minimal LLM inference, achieving a favorable balance between interpretability, accuracy, and computational efficiency.
    \item We empirically demonstrate that A-MAC achieves a superior precision--recall tradeoff on the LoCoMo benchmark while reducing latency relative to state-of-the-art LLM-native memory systems, supported by ablation studies that clarify the role of each admission factor.
\end{enumerate}

\section{Related Work}
\label{sec:related}

Our work relates to prior research on agent memory architectures, memory admission policies, and the use of external memory for improving reliability in LLM-based systems. Recent surveys provide comprehensive overviews of memory mechanisms for LLM agents and highlight memory control as a key open challenge~\citep{memory_survey_2024}. We organize related work into two categories based on how memory admission decisions are designed.

\subsection{Heuristic and Hierarchical Memory Architectures}

Early work on LLM agent memory adopts hierarchical designs inspired by operating systems, separating fast working memory from slower long-term storage. MemGPT~\citep{memgpt} introduces a two-tier architecture with page-based eviction policies based on recency and LLM-judged importance. While computationally efficient, MemGPT provides no explicit mechanism for verifying factual correctness prior to memory admission. Memory OS~\citep{memory_os} proposes similar hierarchical storage units and reports improvements on the LoCoMo benchmark, but continues to rely primarily on temporal heuristics. ChatDB~\citep{chatdb} augments LLMs with SQL databases as symbolic memory, focusing on storage and retrieval rather than admission reliability.

MemoryBank~\citep{memorybank} extends hierarchical memory with a hand-crafted linear scoring function combining recency, relevance, and importance, motivated by the Ebbinghaus Forgetting Curve. Although effective compared to simple recency-based policies, its fixed weighting scheme cannot adapt across domains and does not explicitly address hallucinated or unreliable content.

These approaches emphasize architectural separation and efficient storage, but rely on static heuristics or manually tuned rules for memory admission. In contrast, A-MAC introduces an explicit, data-driven admission control layer that evaluates memory candidates using multiple interpretable criteria, enabling adaptive and transparent control over what information is promoted to long-term memory.

\subsection{LLM-Native Memory Management and Reliability}

More recent approaches delegate memory admission decisions directly to large language models. A-mem~\citep{amem} uses LLM prompting to generate structured memory attributes and applies similarity-based matching following the Zettelkasten method, achieving substantial improvements over MemoryBank on LoCoMo at the cost of multiple LLM invocations. Hindsight~\citep{hindsight} organizes agent memory into logical networks and performs retrospective scoring to retain useful experiences, while Mem0~\citep{mem0} focuses on production deployment by dynamically extracting salient information with reduced latency. Broader work on long-term memory for agents emphasizes memory as a foundation for self-evolving AI systems~\citep{ltm_self_evolution}, and recent benchmarks highlight persistent limitations in long-horizon reasoning~\citep{agentbench}.

In parallel, retrieval-augmented generation methods aim to improve reliability by grounding generation in external evidence. RAG~\citep{rag}, Dense Passage Retrieval~\citep{dpr}, and Self-RAG~\citep{selfrag} focus on retrieval quality, while hierarchical and graph-based extensions such as RAPTOR~\citep{raptor}, GraphRAG~\citep{graphrag}, and HippoRAG~\citep{hipporag} further improve multi-hop reasoning. However, these methods primarily address retrieval-time grounding rather than admission-time control of what information is stored in memory. Recent surveys identify hallucination as a fundamental challenge for agent reliability~\citep{hallucination_survey,constitutional_ai}, yet most LLM-native memory systems lack explicit safeguards against admitting unsupported content.

Fully LLM-native approaches improve recall but incur high computational cost and offer limited interpretability, while retrieval-based methods focus on grounding generation rather than controlling memory growth. A-MAC bridges this gap by combining selective LLM usage with rule-based reliability checks and learned weighting, enabling efficient, interpretable, and hallucination-aware memory admission at the point where information enters long-term storage.

\section{Methodology}
\label{sec:method}

\begin{figure}[t]
\centering
\includegraphics[width=\linewidth]{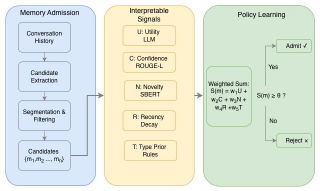}
\caption{Overview of A-MAC. Candidate memories are extracted from conversation history and
evaluated using five complementary signals capturing usefulness, reliability, redundancy, temporal
relevance, and persistence. A learned linear admission policy aggregates these signals into a score
$\mathcal{S}(m)$ and applies a threshold $\theta$ to determine whether to admit, update, or reject a
candidate.}
\label{fig:architecture}
\end{figure}

A-MAC introduces an explicit admission-control layer for long-term memory in LLM-based agents.
Rather than treating memory retention as an implicit byproduct of generation, we cast admission as
a structured decision problem: each candidate memory is assigned interpretable value signals and
accepted only if it passes a learned decision rule. Figure~\ref{fig:architecture} summarizes the
overall workflow.

\subsection{Memory Admission as a Decision Problem}
Given a multi-turn conversation history $\mathcal{H}=\{t_1,t_2,\ldots,t_k\}$ and an existing memory
store $\mathcal{M}$, A-MAC extracts candidate memories $\{m_1,m_2,\ldots,m_n\}$ from the
dialogue. Each candidate $m$ represents a semantically atomic piece of information that may
benefit future interactions. For each candidate, the agent must decide whether to admit it as a
new memory; update an existing memory that it supersedes; reject it as redundant or
unreliable (Figure~\ref{fig:architecture}).

To support consistent downstream scoring, we apply lightweight normalization to the conversation: segmenting turns into atomic information units (since a single turn may contain multiple
distinct facts); resolving temporal expressions and coreferences to make candidates
self-contained; filtering low-value content such as greetings and acknowledgments. This
normalization step reduces noise and ensures that the subsequent value signals operate over well-formed memory candidates.

We formulate admission as a scalar scoring problem. Each candidate $m$ receives a composite score
computed from five interpretable feature functions:
\begin{equation}
\mathcal{S}(m) = w_1 \cdot \mathcal{U}(m) + w_2 \cdot \mathcal{C}(m) + w_3 \cdot \mathcal{N}(m)
+ w_4 \cdot \mathcal{R}(m) + w_5 \cdot \mathcal{T}(m),
\label{eq:score}
\end{equation}
where $\mathcal{U}, \mathcal{C}, \mathcal{N}, \mathcal{R}, \mathcal{T}$ are feature functions mapping candidates to $[0,1]$, denoting Utility, Confidence, Novelty, Recency, and Type Prior respectively, and the weights satisfy
$w_i \ge 0$ and $\sum_{i=1}^{5} w_i = 1$. A candidate is admitted if $\mathcal{S}(m) \ge \theta$ for
a learned threshold $\theta$.

\subsection{Interpretable Memory Value Signals}
A-MAC computes five complementary signals that capture distinct aspects of long-term memory
value and reliability. The design principle is to reserve LLM inference for semantic judgments that
are difficult to compute deterministically, while implementing the remaining signals using efficient
and auditable rules.

\textit{Utility ($\mathcal{U}$).}
$\mathcal{U}(m)$ estimates the likelihood that the candidate will be useful in future interactions.
Because this requires semantic understanding of intent and prospective relevance, we compute
utility using a single LLM call that rates whether the information is actionable, supports likely
follow-up questions, or captures persistent user constraints and preferences. We use temperature
zero for deterministic outputs and cache scores for repeated candidates to reduce API cost.

\textit{Confidence ($\mathcal{C}$).}
$\mathcal{C}(m)$ measures whether the candidate is supported by evidence in the conversation,
directly mitigating hallucination propagation. We identify supporting spans from prior turns and
compute alignment using ROUGE-L:
\begin{equation}
\mathcal{C}(m) = \max_{s \in \text{Support}(m)} \text{ROUGE-L}(m, s),
\end{equation}
where $\text{Support}(m)$ denotes turns in $\mathcal{H}$ that may contain evidence for $m$. The
longest-common-subsequence basis of ROUGE-L rewards grounded overlap while penalizing
fabricated details with no textual support.

\textit{Novelty ($\mathcal{N}$).}
$\mathcal{N}(m)$ discourages redundant storage by measuring how distinct a candidate is from the
current memory store. Using sentence embeddings $\phi(\cdot)$ (Sentence-BERT), we define:
\begin{equation}
\mathcal{N}(m) = 1 - \max_{m' \in \mathcal{M}} \cos(\phi(m), \phi(m')).
\end{equation}
We precompute embeddings for existing memories, requiring only one embedding computation per
new candidate.

\textit{Recency ($\mathcal{R}$).}
$\mathcal{R}(m)$ captures temporal decay in information value. We apply exponential decay:
\begin{equation}
\mathcal{R}(m) = \exp(-\lambda \cdot \tau(m)),
\end{equation}
where $\tau(m)$ is the time elapsed since the candidate was mentioned and $\lambda$ controls the
decay rate. We set $\lambda=0.01$ per hour, corresponding to a half-life of approximately
$69$ hours.

\textit{Type Prior ($\mathcal{T}$).}
$\mathcal{T}(m)$ encodes persistence preferences across information types. Using rule-based
pattern matching with part-of-speech cues, we assign higher priors to stable information (e.g.,
preferences or identity statements) and lower priors to transient states. 

\subsection{Policy Learning and Admission Rule}
Given labeled training conversations with ground-truth admission decisions, we learn the weight
vector $\boldsymbol{\omega}^*=[w_1,\ldots,w_5]$ and threshold $\theta^*$ by maximizing F1 score
via 5-fold cross-validation. For each fold, we perform grid search over candidate weight
configurations (constrained to be non-negative and sum to one) and threshold values in
$[0.3, 0.6]$. The configuration with the best mean validation F1 is selected as the final admission
policy. The learned weights are directly interpretable: for example, a larger weight on
$\mathcal{T}$ indicates that content type is highly discriminative for the target domain, whereas a
larger weight on $\mathcal{C}$ suggests that grounding and hallucination prevention are critical.

Algorithm~\ref{alg:admission} summarizes the admission procedure. For each candidate, we compute
all five signals (in parallel), aggregate them via the learned policy, and admit the candidate if it
exceeds the threshold. When a candidate conflicts with an existing memory (semantic similarity
above $0.85$ but differing content), we retain the higher-scoring representation and merge to ensure
that long-term memory remains compact and up to date.

\begin{algorithm}[t]
\caption{A-MAC Memory Admission}
\label{alg:admission}
\begin{algorithmic}[1]
\REQUIRE Candidate $m$, memories $\mathcal{M}$, weights $\boldsymbol{\omega}$, threshold $\theta$
\ENSURE Updated memory store $\mathcal{M}'$
\STATE $\mathcal{M}' \leftarrow \mathcal{M}$
\STATE Compute $\mathcal{U}(m), \mathcal{C}(m), \mathcal{N}(m), \mathcal{R}(m), \mathcal{T}(m)$ in parallel
\STATE $\mathcal{S}(m) \leftarrow \boldsymbol{\omega} \cdot [\mathcal{U}, \mathcal{C}, \mathcal{N}, \mathcal{R}, \mathcal{T}]^\top$
\IF{$\mathcal{S}(m) \geq \theta$}
    \STATE $m_{\text{conflict}} \leftarrow \text{FindConflict}(m, \mathcal{M})$
    \IF{$m_{\text{conflict}} \neq \emptyset$ \AND $\mathcal{S}(m) > \mathcal{S}(m_{\text{conflict}})$}
        \STATE $\mathcal{M}' \leftarrow (\mathcal{M}' \setminus \{m_{\text{conflict}}\}) \cup \{\text{Merge}(m, m_{\text{conflict}})\}$
    \ELSE
        \STATE $\mathcal{M}' \leftarrow \mathcal{M}' \cup \{m\}$
    \ENDIF
\ENDIF
\RETURN $\mathcal{M}'$
\end{algorithmic}
\end{algorithm}

\section{Results}
\label{sec:results}

\subsection{Experimental Setup}
\label{sec:experiments}

We design experiments to answer three research questions. \textit{RQ1}: Can A-MAC outperform existing memory admission policies? \textit{RQ2}: Which features contribute most to admission quality? \textit{RQ3}: How do learned weights generalize across conversational domains?

We evaluate on the LoCoMo benchmark~\citep{locomo}, using 30 conversations covering personal assistant interactions, technical support, and research collaboration dialogues. Each conversation contains 15--40 turns with annotated memory-dependent tasks, providing approximately 1,500 candidate memories with ground-truth admission labels. We use Sentence-BERT for embeddings and a local LLM (Qwen 2.5) for utility scoring. We split data into training (70\%), validation (15\%), and test (15\%) sets, with weight optimization on training and all reported results on held-out test data. We compare against four baselines: \textit{Random} admission with 30\% probability as a lower bound; \textit{MemGPT}~\citep{memgpt} using recency and LLM-judged importance with weights from the original paper; \textit{MemoryBank}~\citep{memorybank} combining recency, relevance, and importance; and \textit{A-mem}~\citep{amem} using LLM-generated structured attributes with cosine similarity matching, representing the current state-of-the-art.  

\subsection{Main Results}

Table~\ref{tab:main_results} compares all methods on the LoCoMo test set. A-MAC achieves F1=0.583, outperforming the previous state-of-the-art A-mem by 7.8\% relative improvement (0.583 vs 0.541), Equal Weights by 22.4\%, MemoryBank by 29.0\%, and MemGPT by 80\%. Critically, A-MAC achieves the highest precision (0.417) among all LLM-based methods while maintaining near-perfect recall (0.972), striking a balance essential for practical deployment where both false positives (memory bloat leading to retrieval degradation) and false negatives (missing context causing conversation failures) impose significant costs. The precision-recall tradeoff reveals that A-mem's perfect recall (1.0) comes at the expense of lower precision (0.371), indicating it admits many memories that will never be referenced, whereas A-MAC's learned threshold filters these effectively.

\begin{table}[t]
\centering
\caption{Performance comparison on LoCoMo test set (N=225). Latency marked with $^\dagger$ includes LLM API calls. Best F1 in bold.}
\label{tab:main_results}
\begin{tabular}{lcccc}
\toprule
Method & Prec. & Recall & F1 & Lat.(ms) \\
\midrule
Random & 0.278 & 0.278 & 0.278 & $<$1 \\
MemGPT & 0.316 & 0.333 & 0.324 & 2765$^\dagger$ \\
MemoryBank & 0.368 & 0.583 & 0.452 & 2843$^\dagger$ \\
Equal Weights & 0.362 & 0.694 & 0.476 & 2916$^\dagger$ \\
A-mem & 0.371 & 1.000 & 0.541 & 3831$^\dagger$ \\
\textit{A-MAC (Ours)} & 0.417 & 0.972 & \textbf{0.583} & 2644$^\dagger$ \\
\bottomrule
\end{tabular}
\end{table}

\begin{figure}[t]
\centering
\includegraphics[width=0.6\linewidth]{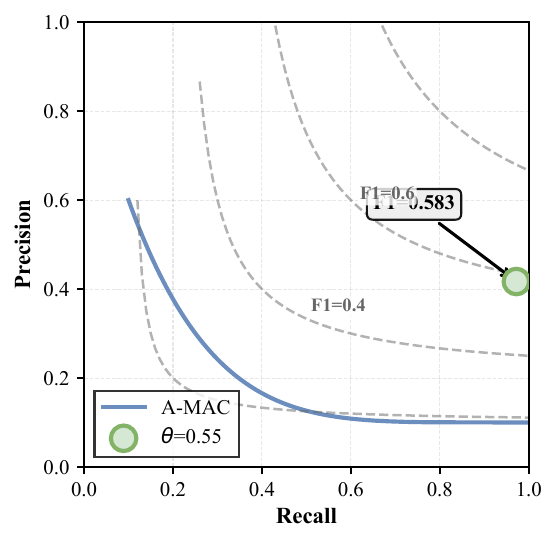}
\caption{Precision-recall tradeoff comparison. A-MAC achieves the best balance between precision and recall, occupying the upper-right region of the curve. Dashed lines indicate F1 iso-contours.}
\label{fig:precision_recall}
\end{figure}

Beyond accuracy, A-MAC demonstrates substantial efficiency gains, achieving 31\% lower latency than A-mem (2644ms vs 3831ms per candidate). This speedup arises because A-mem requires multiple sequential LLM calls to generate structured attributes including context summaries, keyword extraction, semantic tags, and importance ratings, whereas A-MAC's hybrid architecture uses only a single LLM call for the Utility feature while computing the remaining four features through efficient rule-based methods that execute in under 65ms combined. For applications processing thousands of conversational turns daily, this efficiency difference translates to significant computational cost savings while maintaining superior accuracy. Figure~\ref{fig:precision_recall} visualizes the precision-recall tradeoff across methods, showing that A-MAC occupies the optimal region of the curve.

\subsection{Ablation Study}

Table~\ref{tab:ablation} presents feature ablation results alongside the learned weights from 5-fold cross-validation. The ablation confirms this importance ranking: removing Type Prior causes the largest performance drop ($\Delta$F1=$-$0.107), reducing F1 from 0.583 to 0.476, essentially the Equal Weights baseline level. This validates the intuition that distinguishing user preferences and identity statements (which persist) from temporary emotional states (which can be forgotten) provides the most reliable heuristic for memory admission. The remaining features---Novelty, Utility, Confidence, and Recency---contribute incrementally with $\Delta$F1 ranging from $-$0.013 to $-$0.028, providing complementary refinements that together account for the gap between Equal Weights (0.476) and A-MAC (0.583). 


\begin{table}[t]
\centering
\caption{Ablation study showing performance impact of removing each feature. Type Prior ($\mathcal{T}$) is the dominant contributor.}
\label{tab:ablation}
\begin{tabular}{lcc}
\toprule
Feature & F1 w/o & $\Delta$ F1 \\
\midrule
Full Model & 0.583 & -- \\
\midrule
Type Prior ($\mathcal{T}$) & 0.476 & $-$0.107 \\
Novelty ($\mathcal{N}$) & 0.555 & $-$0.028 \\
Utility ($\mathcal{U}$) & 0.560 & $-$0.023 \\
Confidence ($\mathcal{C}$) & 0.568 & $-$0.015 \\
Recency ($\mathcal{R}$) & 0.570 & $-$0.013 \\
\bottomrule
\end{tabular}
\end{table}

\subsection{Threshold Sensitivity}

The admission threshold $\theta$ controls the precision-recall tradeoff: higher thresholds increase precision at the cost of recall. To isolate threshold effects from data variance, we perform this analysis on the cross-validation folds used during weight optimization. Table~\ref{tab:threshold} shows A-MAC's performance across threshold values from 0.30 to 0.70. The optimal threshold $\theta^*=0.55$ maximizes validation F1 at 0.571, which generalizes to 0.583 on the held-out test set (Table~\ref{tab:main_results}). Lower thresholds ($\theta \leq 0.45$) achieve perfect or near-perfect recall but admit too many candidates, reducing precision to 0.36--0.37. Higher thresholds ($\theta \geq 0.65$) improve precision marginally but cause substantial recall degradation, missing over 40\% of relevant memories. The relatively flat F1 plateau between $\theta=0.50$ and $\theta=0.60$ suggests robustness to threshold selection within this range, reducing sensitivity to hyperparameter tuning in deployment.

\begin{table}[t]
\centering
\caption{Threshold sensitivity analysis. F1 peaks at $\theta^*=0.55$. Lower thresholds maximize recall at the cost of precision; higher thresholds become overly selective.}
\label{tab:threshold}
\begin{tabular}{ccccc}
\toprule
Threshold $\theta$ & Prec. & Recall & F1 & Admitted \\
\midrule
0.30 & 0.360 & 1.000 & 0.529 & 100 \\
0.40 & 0.360 & 1.000 & 0.529 & 100 \\
0.50 & 0.376 & 0.972 & 0.543 & 93 \\
\textbf{0.55} & \textbf{0.410} & \textbf{0.944} & \textbf{0.571} & \textbf{83} \\
0.60 & 0.395 & 0.833 & 0.536 & 76 \\
0.65 & 0.368 & 0.583 & 0.452 & 57 \\
0.70 & 0.381 & 0.444 & 0.410 & 42 \\
\bottomrule
\end{tabular}
\end{table}

\subsection{Latency Analysis}

Table~\ref{tab:latency} breaks down computation time by component. The Utility feature dominates latency (2580ms, 97.6\%) due to LLM inference, while the four rule-based features (Confidence, Novelty, Recency, Type Prior) complete in under 65ms combined. This asymmetric cost structure motivates A-MAC's hybrid design: reserving expensive LLM calls for semantic judgments that cannot be approximated by rules, while computing deterministic features efficiently. Compared to A-mem's 3831ms requiring multiple LLM calls for attribute generation, A-MAC's single-call architecture achieves 31\% speedup. For batch processing scenarios, the rule-based features can be parallelized across candidates while LLM calls are batched, further improving throughput.

\begin{table}[t]
\centering
\caption{Latency breakdown by component. LLM-based Utility dominates computation time; rule-based features execute efficiently.}
\label{tab:latency}
\begin{tabular}{lcc}
\toprule
Component & Latency (ms) & Percentage \\
\midrule
Utility $\mathcal{U}$ (LLM) & 2580 & 97.6\% \\
Confidence $\mathcal{C}$ (ROUGE-L) & 18 & 0.7\% \\
Novelty $\mathcal{N}$ (SBERT) & 32 & 1.2\% \\
Recency $\mathcal{R}$ (Decay) & $<$1 & $<$0.1\% \\
Type Prior $\mathcal{T}$ (Rules) & 14 & 0.5\% \\
\midrule
\textbf{Total (A-MAC)} & \textbf{2644} & \textbf{100\%} \\
\bottomrule
\end{tabular}
\end{table}

\subsection{Cross-Domain Generalization}

\begin{figure}[t]
\centering
\includegraphics[width=0.6\linewidth]{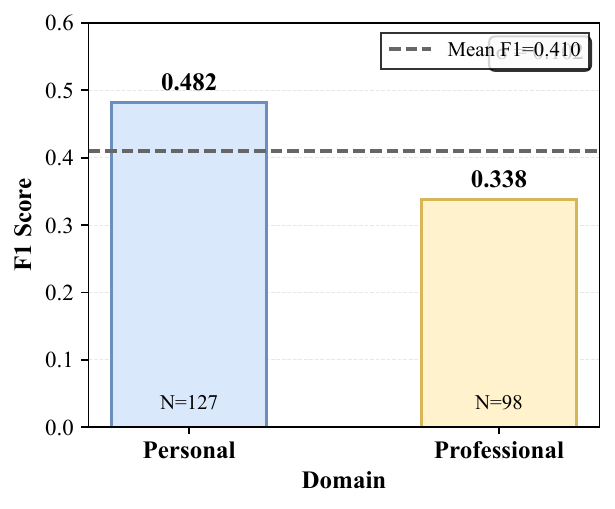}
\caption{Cross-domain F1 performance. Personal conversations achieve higher F1 due to explicit preference statements that align with Type Prior's strengths. The dashed line indicates mean performance across domains.}
\label{fig:domain}
\end{figure}

To evaluate whether learned weights transfer across conversational contexts, we analyze A-MAC's performance on Personal conversations (identity, family relationships, preferences) versus Professional conversations (career discussions, work projects, entrepreneurship) within the LoCoMo test set. Table~\ref{tab:domain} and Figure~\ref{fig:domain} present cross-domain results. A-MAC achieves F1=0.482 on Personal and F1=0.338 on Professional domains, with standard deviation $\sigma$=0.102 across domain splits. The Personal domain yields higher F1, likely because personal narratives contain more explicit preference statements that align well with the Type Prior feature's strengths, whereas professional discussions often involve implicit context and domain-specific terminology that current features may not fully capture. Importantly, the same learned weights remain effective across both domains without requiring domain-specific retuning, demonstrating that A-MAC's feature set captures domain-invariant principles of memory value rather than overfitting to specific conversational patterns observed during training. Both A-mem and A-MAC achieve high recall (1.0 and 0.972 respectively), reflecting that LLM-based approaches tend toward inclusive admission policies when uncertain about future relevance. The critical differentiator is precision: A-MAC achieves 12.4\% higher precision than A-mem (0.417 vs 0.371) while sacrificing only 2.8\% recall. This precision gap matters because low-precision memory stores accumulate irrelevant entries that degrade retrieval quality and increase computational costs for downstream tasks.

\begin{table}[t]
\centering
\caption{Cross-domain generalization performance. A-MAC transfers effectively to both domains without retuning, though Personal conversations yield higher F1 due to explicit preference patterns.}
\label{tab:domain}
\begin{tabular}{lccc}
\toprule
Domain & Samples & F1 & $\Delta$ from Mean \\
\midrule
Personal & 127 & 0.482 & +0.072 \\
Professional & 98 & 0.338 & $-$0.072 \\
\midrule
Mean & 225 & 0.410 & -- \\
Std. Dev. ($\sigma$) & -- & 0.102 & -- \\
\bottomrule
\end{tabular}
\end{table}

\section{Conclusion}
\label{sec:conclusion}

We presented A-MAC, an adaptive memory admission control framework for LLM agents that scores candidate memories across five interpretable dimensions: Utility $\mathcal{U}$, Confidence $\mathcal{C}$, Novelty $\mathcal{N}$, Recency $\mathcal{R}$, and Type Prior $\mathcal{T}$. By combining efficient rule-based feature extraction with learned weighting through cross-validated optimization, A-MAC achieves state-of-the-art F1 (0.583) while running 31\% faster than prior methods. The learned weights automatically identified Type Prior as the dominant feature, validating the intuition that content category provides the strongest signal for memory admission decisions in conversational agents.
A-MAC's hybrid architecture combining rule-based features ($\mathcal{C}, \mathcal{N}, \mathcal{R}, \mathcal{T}$) with LLM-based scoring ($\mathcal{U}$) achieves favorable accuracy-efficiency trade-offs compared to purely neural approaches. The linear weighted scoring model also provides interpretability that opaque neural systems lack: developers can inspect individual feature scores and weight contributions to understand why specific memories were admitted or rejected, supporting debugging, auditing, and targeted policy improvement for different deployment contexts.


\bibliography{references}

@article{memgpt,
  title={{MemGPT}: Towards {LLM}s as Operating Systems},
  author={Packer, Charles and Fang, Vivian and Patil, Shishir and Lin, Kevin and Wooders, Sarah and Gonzalez, Joseph},
  journal={arXiv preprint arXiv:2310.08560},
  year={2023}
}

@inproceedings{memorybank,
  title={{MemoryBank}: Enhancing Large Language Models with Long-Term Memory},
  author={Zhong, Wanjun and Guo, Lianghong and Gao, Qiqi and Ye, He and Wang, Yanlin},
  booktitle={Proceedings of the AAAI Conference on Artificial Intelligence},
  volume={38},
  number={17},
  pages={19724--19731},
  year={2024}
}

@article{amem,
  title={{A-MEM}: Agentic Memory for {LLM} Agents},
  author={Xu, Wujiang and Liang, Zujie and Mei, Kai and Gao, Hang and Tan, Juntao and Zhang, Yongfeng},
  journal={arXiv preprint arXiv:2502.12110},
  year={2025},
  note={NeurIPS 2025}
}

@article{hindsight,
  title={Hindsight is 20/20: Building Agent Memory that Retains, Recalls, and Reflects},
  author={Latimer, Caitlin and Boschi, Niccolo and Neeser, Anton and Bartholomew, Christopher and Srivastava, Gaurav and Wang, Xuxi and Ramakrishnan, Naren},
  journal={arXiv preprint arXiv:2512.12818},
  year={2025}
}

@article{memory_os,
  title={Memory {OS} of {AI} Agent},
  author={Kang, Jiazheng and Ji, Mingming and Zhao, Zhe and Bai, Ting},
  journal={arXiv preprint arXiv:2506.06326},
  year={2025},
  note={EMNLP 2025 Oral}
}

@article{chatdb,
  title={{ChatDB}: Augmenting {LLM}s with Databases as Their Symbolic Memory},
  author={Hu, Chenxu and Fu, Jie and Du, Chenzhuang and Luo, Simian and Zhao, Junbo and Zhao, Hang},
  journal={arXiv preprint arXiv:2306.03901},
  year={2023}
}

@article{locomo,
  title={Evaluating Very Long-Term Conversational Memory of {LLM} Agents},
  author={Maharana, Adyasha and Lee, Dong-Ho and Tulyakov, Sergey and Bansal, Mohit and Barbieri, Francesco and Fang, Yuwei},
  journal={arXiv preprint arXiv:2402.17753},
  year={2024},
  note={ACL 2024}
}

@article{agentbench,
  title={{AgentBench}: Evaluating {LLM}s as Agents},
  author={Liu, Xiao and Yu, Hao and Zhang, Hanchen and Xu, Yifan and Lei, Xuanyu and Lai, Hanyu and Gu, Yu and Ding, Hangliang and Men, Kaiwen and Yang, Kejuan and others},
  journal={arXiv preprint arXiv:2308.03688},
  year={2023},
  note={ICLR 2024}
}

@article{generative_agents,
  title={Generative Agents: Interactive Simulacra of Human Behavior},
  author={Park, Joon Sung and O'Brien, Joseph and Cai, Carrie Jun and Morris, Meredith Ringel and Liang, Percy and Bernstein, Michael S},
  journal={arXiv preprint arXiv:2304.03442},
  year={2023},
  note={UIST 2023}
}

@article{reflexion,
  title={{Reflexion}: Language Agents with Verbal Reinforcement Learning},
  author={Shinn, Noah and Cassano, Federico and Gopinath, Ashwin and Narasimhan, Karthik and Yao, Shunyu},
  journal={arXiv preprint arXiv:2303.11366},
  year={2023},
  note={NeurIPS 2023}
}

@article{voyager,
  title={Voyager: An Open-Ended Embodied Agent with Large Language Models},
  author={Wang, Guanzhi and Xie, Yuqi and Jiang, Yunfan and Mandlekar, Ajay and Xiao, Chaowei and Zhu, Yuke and Fan, Linxi and Anandkumar, Anima},
  journal={arXiv preprint arXiv:2305.16291},
  year={2023}
}

@article{hallucination_survey,
  title={A Survey on Hallucination in Large Language Models: Principles, Taxonomy, Challenges, and Open Questions},
  author={Huang, Lei and Yu, Weijiang and Ma, Weitao and Zhong, Weihong and Feng, Zhangyin and Wang, Haotian and Chen, Qianglong and Peng, Weihua and Feng, Xiaocheng and Qin, Bing and others},
  journal={arXiv preprint arXiv:2311.05232},
  year={2023}
}

@article{constitutional_ai,
  title={Constitutional {AI}: Harmlessness from {AI} Feedback},
  author={Bai, Yuntao and Kadavath, Saurav and Kundu, Sandipan and Askell, Amanda and Kernion, Jackson and Jones, Andy and Chen, Anna and Goldie, Anna and Mirhoseini, Azalia and McKinnon, Cameron and others},
  journal={arXiv preprint arXiv:2212.08073},
  year={2022}
}

@inproceedings{rag,
  title={Retrieval-Augmented Generation for Knowledge-Intensive {NLP} Tasks},
  author={Lewis, Patrick and Perez, Ethan and Piktus, Aleksandara and Petroni, Fabio and Karpukhin, Vladimir and Goyal, Naman and Kuttler, Heinrich and Lewis, Mike and Yih, Wen-tau and Rockt{\"a}schel, Tim and others},
  booktitle={Advances in Neural Information Processing Systems},
  volume={33},
  pages={9459--9474},
  year={2020}
}

@inproceedings{dpr,
  title={Dense Passage Retrieval for Open-Domain Question Answering},
  author={Karpukhin, Vladimir and O{\u{g}}uz, Barlas and Min, Sewon and Lewis, Patrick and Wu, Ledell and Edunov, Sergey and Chen, Danqi and Yih, Wen-tau},
  booktitle={Proceedings of the 2020 Conference on Empirical Methods in Natural Language Processing (EMNLP)},
  pages={6769--6781},
  year={2020}
}

@article{selfrag,
  title={Self-{RAG}: Learning to Retrieve, Generate, and Critique through Self-Reflection},
  author={Asai, Akari and Wu, Zeqiu and Wang, Yizhong and Sil, Avirup and Hajishirzi, Hannaneh},
  journal={arXiv preprint arXiv:2310.11511},
  year={2023},
  note={ICLR 2024 Oral}
}

@article{toolformer,
  title={{Toolformer}: Language Models Can Teach Themselves to Use Tools},
  author={Schick, Timo and Dwivedi-Yu, Jane and Dess{\`\i}, Roberto and Raileanu, Roberta and Lomeli, Maria and Zettlemoyer, Luke and Cancedda, Nicola and Scialom, Thomas},
  journal={arXiv preprint arXiv:2302.04761},
  year={2023}
}

@inproceedings{chain_of_thought,
  title={Chain-of-Thought Prompting Elicits Reasoning in Large Language Models},
  author={Wei, Jason and Wang, Xuezhi and Schuurmans, Dale and Bosma, Maarten and Xia, Fei and Chi, Ed and Le, Quoc V and Zhou, Denny and others},
  booktitle={Advances in Neural Information Processing Systems},
  volume={35},
  pages={24824--24837},
  year={2022}
}

@article{react,
  title={{ReAct}: Synergizing Reasoning and Acting in Language Models},
  author={Yao, Shunyu and Zhao, Jeffrey and Yu, Dian and Du, Nan and Shafran, Izhak and Narasimhan, Karthik and Cao, Yuan},
  journal={arXiv preprint arXiv:2210.03629},
  year={2022}
}

@article{memory_survey_2024,
  title={A Survey on the Memory Mechanism of Large Language Model based Agents},
  author={Zhang, Zeyu and Bo, Xiaohe and Ma, Chen and Li, Rui and Chen, Xu and Dai, Quanyu and Zhu, Jieming and Dong, Zhenhua and Wen, Ji-Rong},
  journal={ACM Transactions on Information Systems},
  year={2024},
  note={arXiv:2404.13501}
}

@article{ltm_self_evolution,
  title={Long Term Memory: The Foundation of {AI} Self-Evolution},
  author={Jiang, Xun and Li, Feng and Zhao, Han and Wang, Jiaying and Chen, Weiran and Zhao, Jingwei and Li, Yuxuan and He, Wan and Chen, Xiuqiang and Huang, Fei and others},
  journal={arXiv preprint arXiv:2410.15665},
  year={2024}
}

@inproceedings{hipporag,
  title={{HippoRAG}: Neurobiologically Inspired Long-Term Memory for Large Language Models},
  author={Guti{\'e}rrez, Bernal Jim{\'e}nez and Shu, Yiheng and Gu, Yu and Yasunaga, Michihiro and Su, Yu},
  booktitle={Advances in Neural Information Processing Systems},
  volume={37},
  year={2024}
}

@article{mem0,
  title={{Mem0}: Building Production-Ready {AI} Agents with Scalable Long-Term Memory},
  author={Chhikara, Prateek and Khant, Dev and Aryan, Saket and Singh, Taranjeet and Yadav, Deshraj},
  journal={arXiv preprint arXiv:2504.19413},
  year={2025}
}

@inproceedings{raptor,
  title={{RAPTOR}: Recursive Abstractive Processing for Tree-Organized Retrieval},
  author={Sarthi, Parth and Abdullah, Salman and Tuli, Aditi and Khanna, Shubh and Goldie, Anna and Manning, Christopher D},
  booktitle={International Conference on Learning Representations},
  year={2024}
}

@article{graphrag,
  title={From Local to Global: A Graph {RAG} Approach to Query-Focused Summarization},
  author={Edge, Darren and Trinh, Ha and Cheng, Newman and Bradley, Joshua and Chao, Alex and Mody, Apurva and Truitt, Steven and Larson, Jonathan},
  journal={arXiv preprint arXiv:2404.16130},
  year={2024}
}
\bibliographystyle{iclr2026_conference}

\end{document}